\documentclass[letterpaper, 10 pt, conference]{ieeeconf}  

\IEEEoverridecommandlockouts                              

\usepackage{cite}
\usepackage{amsmath,amssymb,amsfonts}
\usepackage{algorithmic}
\usepackage{graphicx}
\usepackage{textcomp}
\usepackage{xcolor}
\usepackage{tikz}
\usepackage{pgfplots}
\usepackage{subcaption}
\usepackage{booktabs}
\usepackage{adjustbox}
\usepackage{rotating}
\usepackage{geometry}
\begin{document}

\title{Enhancing the Performance of Pneu-net Actuators Using a Torsion Resistant Strain Limiting Layer} 

\author{Ian Good$^{1}$ Srivatsan Balaji$^{1}$ and Jeffrey I Lipton$^{2*}$
\thanks{This work was supported by the National Science Foundation, grant numbers 2017927 and 2035717, by the ONR through Grant DB2240 and by the Murdock Charitable Trust through grant 201913596}
\thanks{$^{1}$ Mechanical Engineering Department of the University of Washington, Seattle, WA, 98195 USA }%
\thanks{$^{2}$ Mechanical and Industrial Engineering Department of Northeastern University, Boston, MA, 02115 USA }%
\thanks{$^{*}$ {\tt\small j.lipton@northeastern.edu}}%
}

\maketitle
\thispagestyle{empty}
\pagestyle{empty}

\begin{abstract}
Pneunets are the primary form of soft robotic grippers.
A key limitation to their wider adoption is their inability to grasp larger payloads due to objects slipping out of grasps. 
We have overcome this limitation by introducing a torsionally rigid strain limiting layer (TRL).
This reduces out-of-plane bending while maintaining the gripper's softness and in-plane flexibility. 
We characterize the design space of the strain limiting layer for a Pneu-net gripper using simulation and experiment and map bending angle and relative grip strength.
We found that the use of our TRL reduced out-of-plane bending by up to 97.7\% in testing compared to a benchmark Pneu-net gripper from the Soft Robotics Toolkit. We demonstrate a lifting capacity of 5kg when loading using the TRL. We also see a relative improvement in peak grip force of 3N and stiffness of 1200N/m compared to 1N and 150N/m for a Pneu-net gripper without our TRL at equal pressures.
Finally, we test the TRL gripper on a suite of six YCB objects above the demonstrated capability of a traditional Pneu-net gripper. We show success on all but one demonstrating significant increased capabilities.

\end{abstract}


\section{Introduction}

\begin{figure}[t]
    \centering
    \includegraphics[width=0.98\linewidth]{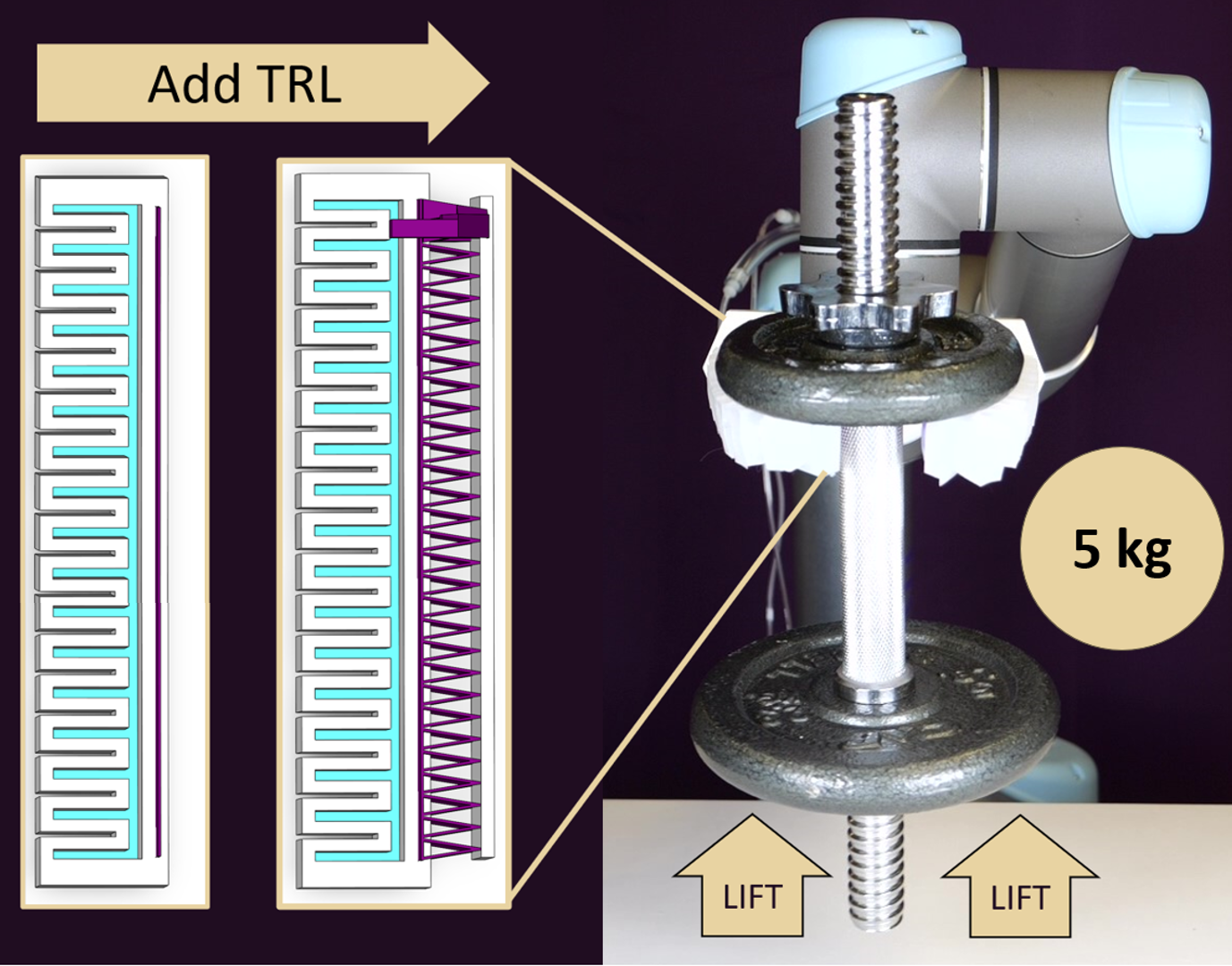}
    \caption{We create a new Torsion Resistant Strain Limiting Layer (TRL) that can be added to any existing Pneu-net based gripper. It increases their resistance to torsion, allowing the gripper to lift larger payloads. Additionally, we can load the skeleton of the TRL directly, dramatically increasing lifting capacity. We show the TRL gripper lifting a 5kg dumbbell using the skeleton, the maximum payload capacity of the UR5 robot arm.}
    \label{fig:hero}
\end{figure}

Out-of-plane deformations limit a soft robotic grippers' ability to grasp and manipulate heavy objects. Soft robotic systems have been able to achieve large payload capacities, able to lift car tires and dumbbells\cite{li2017fluid,li2019vacuum,li2022scaling}. However, this has required the gripper to cage the object or pull directly against gravity. Holding objects perpendicular to gravity or having large movements cause soft robotic grippers to deform and twist and ultimately loose their grasp. A second limitation of soft grippers has been that the out-of-plane deformation in the fingers leads to uncertainty in the positioning of the grasped object. A successful but heavy grasp can cause a gripper to deform in such a way that localizing relative to the arm is difficult. Here we show how a Torsion Resistant layer can increase the out-of-plane payload capacity of Pneu-net grippers to 5kg (Fig\ref{fig:hero}).

For Pneu-nets\cite{mosadegh2014pneumatic}, this fundamental limitation comes from the use of simple rectangular beams as the Strain Limiting Layer. Rectangular cross sections have low second moments of inertia in one direction, allowing them to bend in the thin direction. While in theory this would make them able to resist deformations along their thick direction, in practice these structures have a low resistance to torsion. This causes any out-of-plane loading to generate a twist in the Strain Limiting Layer (SLL).

Several strategies have emerged for improving the out-of-plane performance of soft grippers through the SLL. One approach focuses on changing the material distribution by using multi-material topology optimization\cite{wang2022topology}. Another approach discritizes the SLL and inserts hinges to allow a thicker material to still bend \cite{lotfiani2020torsional}. The first still experiences large out of plane deformations while demonstrating smaller increases in torsional stiffness. The second requires significantly higher input pressures and the inclusion of hinge based Strain Limiting Layers (SLL) can reduce the structures ability to conform to objects.

We focus on using advancements from the field of compliant mechanisms to change the geometry of the SLL. We use triangulated beams as the Torsion Resistant Strain Limiting Layer (TRL) of Pneu-net. These triangulated beams resist torsion but allow bending on planar and spherical surfaces \cite{rommers2021new,naves2017building}. While one might expect the triangles along the beam to localize the deformation to interfaces between triangles, the structures deform over their entire length, with all sides of the triangle deforming when bent. The triangularized beams act as a continuum structure with enhanced stiffness to out of plane bending and torsion. 

We integrate the TRLs into standard Pneu-net grippers and provide a step-by-step guide for integrating the TRLs. Next, we perform a parametric sweep on the TRLs to provide a design guide. We found that there was minimal difference between triangle designs for in-plane bending compared to flat Strain Limiting Layers. We found a trade-off between torsional stiffness and stress concentrations in the TRL. We tested the integrated gripper on a cylindrical pull test and found that the addition of the TRL increased the gripping force by 3x and increase the stiffness in the grip by 7.8x. Finally, we tested the gripper by picking up a weighted structure entirely using the sides of the gripper. As seen in Figure 1, the TRL gripper can pick up a 5kg weight (the max capacity of our UR5) using lateral loading.

In this paper we:
\begin{itemize}
    \item Model key parameters for understanding a Torsion Resistant Strain Limiting Layer
    \item Characterize in and out of plane bending performance against a benchmark Pneu-net gripper 
    \item Demonstrate the increased payload and lifting capacity of the Torsion Resistant Strain Limiting Layer
\end{itemize}

\section{Background}
Soft robotics researchers have attempted to improve the out-of-plane performance of Pneu-nets using several techniques. 
Some look to redesign the gripper as a whole. This can involve  putting a skeleton around the entire gripper\cite{scharff2019reducing}. However, this gripper required significantly more input pressure to reach the same normal force output and demonstrates smaller lifting capacity than the gripper presented here. Another method involves optimizing the design of individual cells to improve torsion resistance\cite{su2022optimizing}. It demonstrates less stiffness when normalizing for length.

Active Strain Limiting Layers are another area of active research. Many of these change their SLL stiffness through phase change \cite{zhang2019fast}, jamming chain, \cite{jiang2019chain} or through the thermoelectric properties of Field's metal \cite{buckner2019enhanced,gunawardane2022thermoelastic} or by heating hydrogels, causing them to swell\cite{visentin2021selective}. Additionally, work has been done to use jamming to increase a grippers ability to grasp objects and increase out of plane grip strength\cite{crowley20223d}. Many of these require significant infrastructure to support their use relative to a passive Pneu-net actuator, and are more challenging to manufacture.

Our solution allows a simple passive SLL made from a single material to be easily integrated into existing Pneu-net designs. It demonstrates a large torsional spring constant and high normal force output for a comparatively low input pressure. It can be made with a low cost 3D printer and uses no additional infrastructure beyond a benchmark Pneu-net.

\section{Derivation of the effect of Torsion on Antipodal Gripping for Soft Bodies}
\begin{figure}
    \centering
    \includegraphics[width=0.98\linewidth]{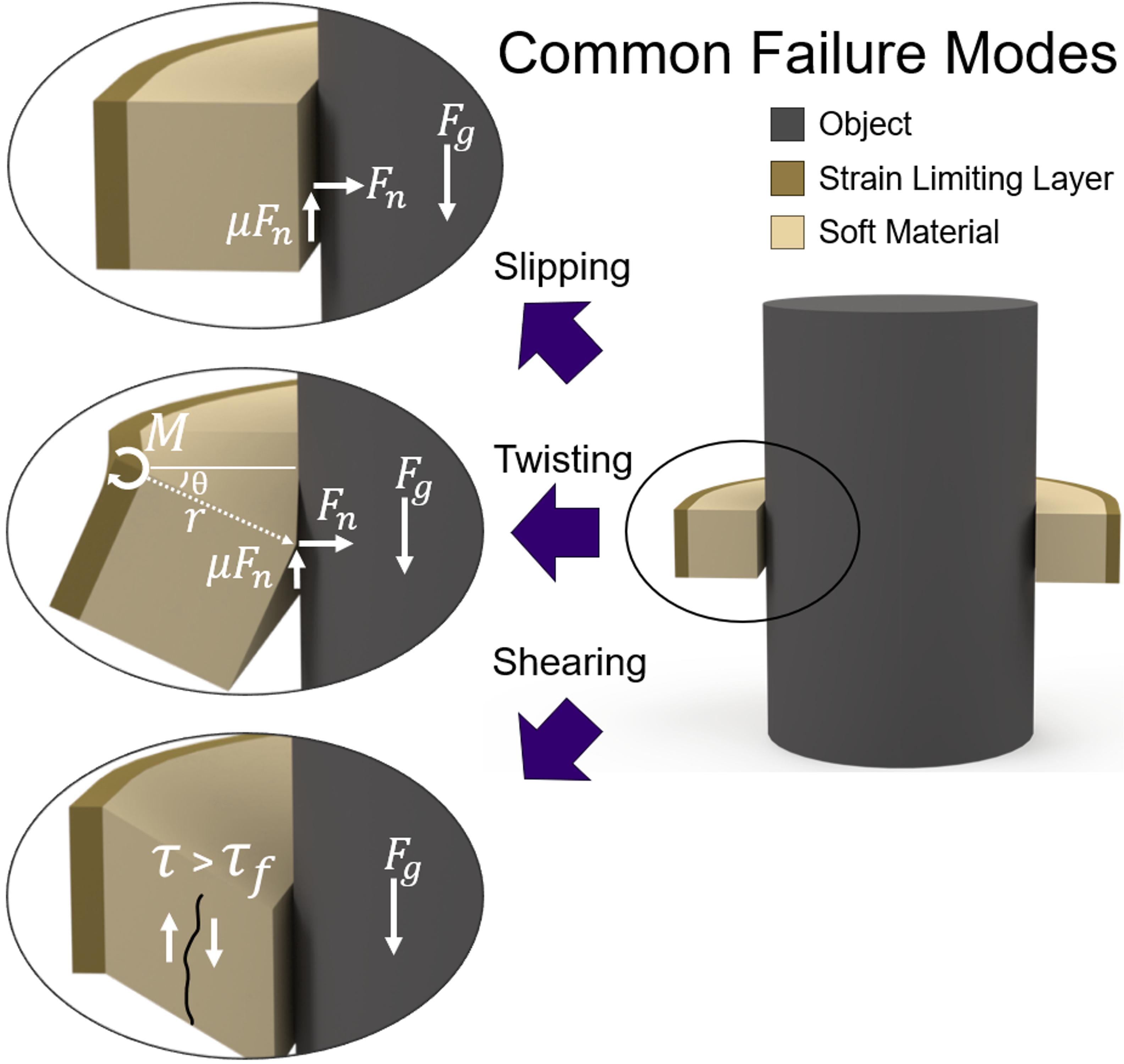}
    \caption{We demonstrate the common failure modes for Pneu-net grasping. On the right we can see a simplified gripper grabbing a cylinder using an antipodal grasp. Only the Strain Limiting Layer and soft gripping material are shown. On the left, three common failure modes are shown; slipping due to low normal forces, twisting due to torsional deformation in the SLL, and shearing of the soft material. The maximum payload capacity for a Pneu-net gripper is governed by the minimum of the set of these. Our gripper increases the torsional resistance and normal force applied compared to a standard Pneu-net gripper. This allows our gripper to lift larger payloads while having a better understanding of where the object is in our grasp.}
    \label{fig:failureModes}
\end{figure}

Antipodal gripping is a widespread application for soft robotic grippers. It involves picking up an object through two opposite points. There are three common limits to a soft grippers ability to perform an antipodal grasp; insufficient normal force, slippage due to torsional deformation, and shearing of the soft layer. All three of these modes as well as an example antipodal grasp can be seen in Fig. \ref{fig:failureModes}. The minimum of these three modes govern the payload capacity for a simply lifted object. If the payload minimum is governed by insufficent normal force, the force a gripper can lift is shown in Eq. \ref{eq:staticForce}.

\begin{equation}
     F_{static} = \mu F_n
     \label{eq:staticForce}
\end{equation}

where $F_{static}$ is the maximum force a gripper can lift, $\mu$ is the coefficient of static friction, and $F_n$ is the normal force the gripper applies as shown in Fig. \ref{fig:failureModes}. If instead the gripper is just at the limit of slip from twisting, we get:

\begin{equation}
     M = {F_gr}
     \label{eq:staticTorque}
\end{equation}

\begin{equation}
     M = \kappa\theta
     \label{eq:Torque}
\end{equation}

where $M$ is the Torque on the Strain Limiting Layer, $F_g$ is the gravitational force from the object, $r$ is the distance from the neutral axis of the gripper to the object, $\kappa$ is the torsional spring constant, and $\theta$ is the angle of twist relative to the normal force. These can be seen labeled in the twisting failure mode in Fig. \ref{fig:failureModes}. If we assume there is some small vertical deformation $x$ as the gripper is loaded but no slip or rolling occur, we get Eq. \ref{eq:xsmallAng}.

\begin{equation}
     x = r\theta
     \label{eq:xsmallAng}
\end{equation}

Combining Eq. \ref{eq:staticTorque}, Eq. \ref{eq:Torque}, and Eq. \ref{eq:xsmallAng} we can solve for $F_g$ as a function of $x$:

\begin{equation}
    F_g = \dfrac{\kappa}{r^2} x
    \label{eq:forceG}
\end{equation}

Where $\dfrac{\kappa}{r^2}$ is the effective stiffness of the system.

Shearing of the soft gripping material can be another concern and is defined when:
\begin{equation}
     \tau > \tau_{f} 
     \label{eq:shearStress}
\end{equation}

Where $\tau$ is the length of the shear vector in the material and $\tau_{f}$ is the shear fracture of the material. For this study $\tau$ was never found to be greater than $\tau_{f}$. 

Together the minimum force value from Eq. \ref{eq:staticForce}, Eq. \ref{eq:forceG}, and Eq. \ref{eq:shearStress} govern the payload capacity of a soft gripper performing an anti-podal pick on a simple object. Since Pneu-nets grippers often demonstrate out of plane deformation as the most common failure mode, this work looks to Eq. \ref{eq:forceG} as a way to increase the payload capacity of Pneu-net grippers.


\section{Design of Torsion Resistant Strain Limiting Layer}

This section establishes a benchmark of comparison for our TRL. It explores the design space for Strain Limiting Layers and evaluates them based on their in-plane and angular displacement as a function of thickness. We explore the design space of the TRL gripper in simulation by varying the number of triangles present over the length from two to thirty. Finally, we use the simulation results to inform which gripper to instantiate in the real world. We present the fabrication process and test methods for comparison for the TRL compared to the SLL. Based on the analysis in section A, a strong torsional spring constant will be critical for a high payload capacity. Additionally, strong in-plane bending performance will allow the gripper to lift a larger range of objects for a given antipodal mounting distance.

\subsection{Benchmark Design}
We used the Pneu-net from the Soft Robotics Toolkit as our benchmark design. \cite{holland2014soft}. We create our TRL gripper by replacing the benchmark SLL with a 3D-printed TRL and adding a thin layer of Ecoflex 00-30 to the tips of the triangles to maintain the same gripping surface. Both the benchmark and TRL gripper can be seen in cross section in Fig.  \ref{fig:hero} with the air pockets highlighted in blue, and the Strain Limiting Layers highlighted in purple. 

\subsection{Traditional Strain Limiting Layers}
We need to establish a baseline of comparison for our TRL compared to a traditional Strain Limiting Layer. We chose a rectangular beam with a fixed length of 100 mm and a width of 20 mm was used as the base of the SLL as this matched the 2-D projection of the elastomer pockets in a Pneu-net. The only free parameter in the design is the number of triangles and base material. We chose PA-6 as the base material because it is significantly stiffer than Ecoflex 00-30, has favorable elasticity properties, is easy to manufacture into complex shapes, and had similar properties as PLA which was used to fabricate the TRL.

To understand the in-plane and torsional performance of a standard SLL, we analyzed the design using finite element analysis (FEA). We varied the thickness of the SLL and compared the in-plane bending and angular displacement. In-plane deformations are important as they allow a large range of objects to be grasped and can be used to show the energy efficiency of a design. For a given pressure, a design that bends more in the plane will use less energy to deform the structure, resulting in larger normal forces applied. Torsional deflection is important because a major failure mode for soft gripper not being able to lift an object is due to twisting of the grasp. Additionally, reduced torsional deflection allows for a better understanding of where objects are in the grasp.

The data from the FEA can see seen in Table \ref{tab:ssl-analysis}. We can see increases in in-plane and angular displacement as the SLL thickness is decreased. The performance ratio of in-plane displacement over angular displacement also improves as thickness decreases. Here we apply a moment that is an order of magnitude smaller than the moment applied to the TRL. This is because applying the same moment resulted in non-real geometry for the SLL.

\begin{table}[t]
\caption{Simulated In-Plane Deformation and Angular Displacement for Simple Strain Limiting Layer (SLL). }
\begin{tabular}{ccc}
\hline
\begin{tabular}[c]{@{}c@{}}Thickness\\ {[}mm{]}\end{tabular} & \begin{tabular}[c]{@{}c@{}}In-Plane Displacement \\ (F = 0.01 N)  {[}mm{] }\end{tabular} & \begin{tabular}[c]{@{}c@{}}Angular Displacement \\ (M = 0.5 N mm)  {[}degrees{]}\end{tabular} \\ \hline
0.3 & 64.85 & 32.89 \\
0.4 & 27.38 & 15.36 \\
0.5 & 14.02 & 8.02 \\
0.6 & 8.11 & 4.70 \\
0.7 & 5.11 & 2.98 \\
0.8 & 3.42 & 2.01 \\
0.9 & 2.41 & 1.43 \\
1.0 & 1.75 & 1.03 \\ \hline
\end{tabular}
\label{tab:ssl-analysis}
\end{table}

\subsection{Simulation of Torsion Resistant Layer (TRL)}
\begin{figure}[t]
    \centering
    \includegraphics[width=0.45\textwidth]{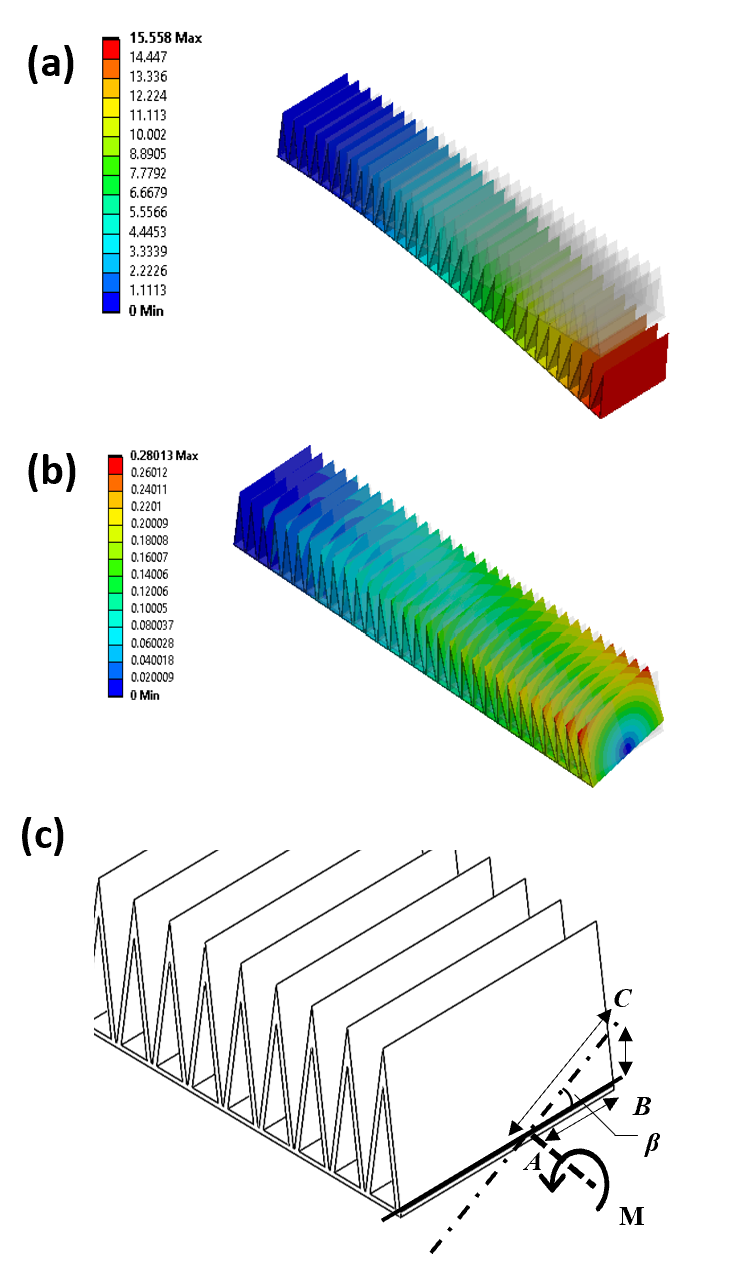}
    \caption{FEA results of (a) in-plane bending simulation, (b) torsion test and (c) torsional bending setup for simulated angle measurement. A moment M is applied along the longitudinal axis of the TRL. Line AB is deformed to line AC forming the angle, $\beta$, which is calculated using Eq. \ref{eq:angle-calculation}. The number of triangles was varied across a span of 100mm resulting in different triangle widths.}
    \label{fig:angle}
\end{figure}

Our first step was to analyze the design space of the TRL. 
The main design parameter for the TRL was the angle $\alpha$ between the base and the side of the triangle. We fixed the length of the TRL to be the length of the SLL (100mm) from the reference design. Because our design relies on having complete triangles, this discritizes our search space to be angles that result in integer numbers of triangles along the length of the TRL. The number of triangles varied from two to thirty. The thirty triangle configuration can be seen in Fig.  \ref{fig:angle}. We fixed the height of each triangle to be 10\% the major dimension, plus 2mm for bonding a gripping layer to the end of the TRL. This was chosen as a reasonable value to search the design space without significantly increasing the volume of the gripper. By keeping this value small, we enable easy retrofitting onto grippers already using Pneu-nets.

In-plane and torsional bending conditions were simulated using Ansys Mechanical 2022. PA-6 was used as the model material as its properties are similar to the PLA used in the printed TRL. For the bending test, one end of the skeleton was fixed and an in-plane bending force of 0.01N was applied to the other edge of the beam. This roughly simulates the condition that occurs when the air chambers are pressurized.

\begin{figure}[t]
    \centering
    \includegraphics[width=.45\textwidth]{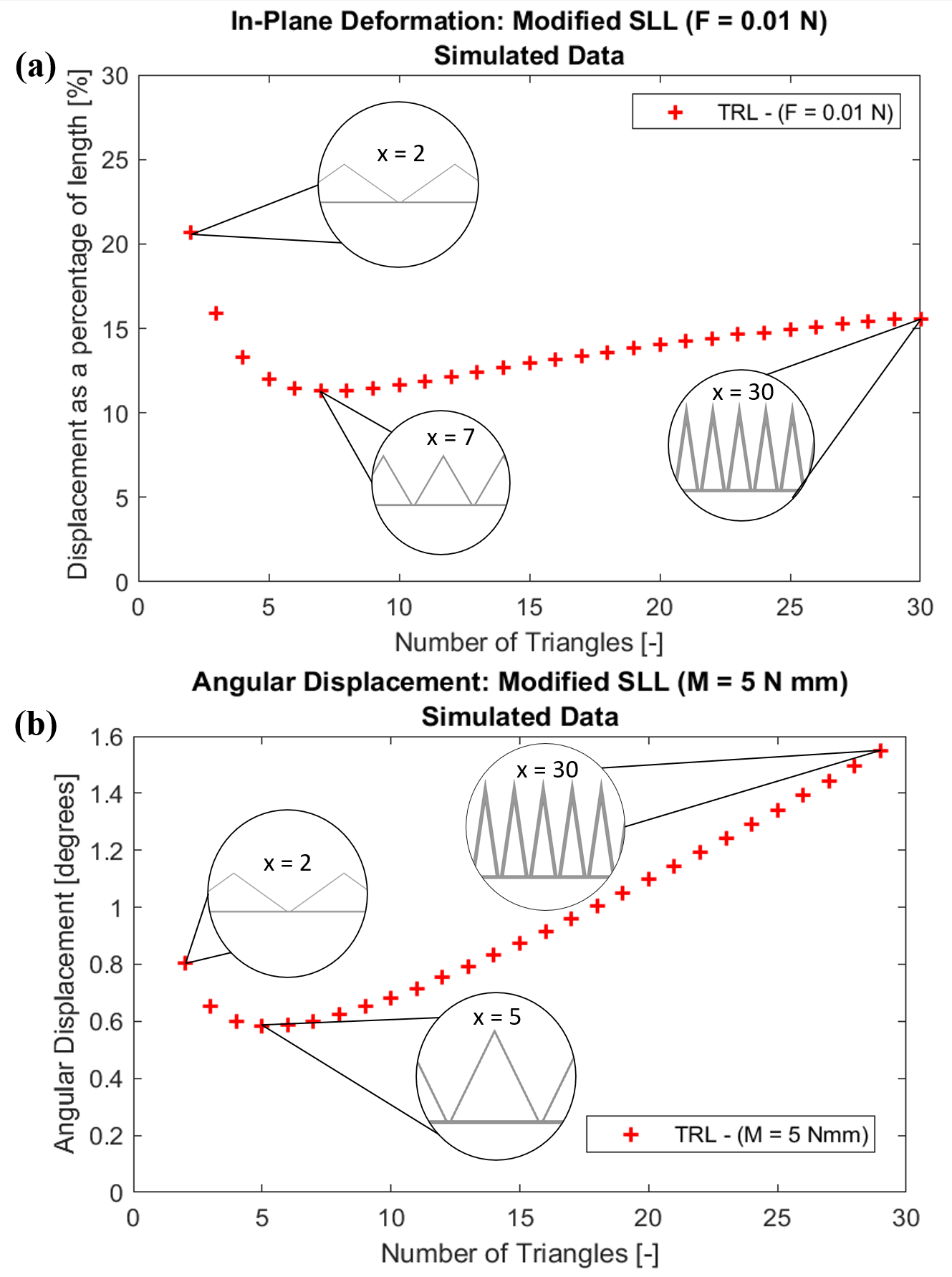}
    \caption{Plot of In-plane displacement (a) and Angular displacement (b) as a function of number of triangles in the Strain Limiting Layer. In (a), we see in-plane deformations from an applied load of 0.01N and in (b), we see the angular displacement from a 5Nmm torque. Both plots follow a swoosh pattern with local maxima at the extreme ends of the range (two and thirty triangles respectively). We choose the 30 triangle solution for our TRL gripper since there is little difference in the angular displacement compared to the increase from adding the triangulated elements to the SLL. This choice reduces the stress concentration within the SLL, expanding lifetime of the gripper.}
    \label{fig:triangleFEAresults}
    \vspace{-0.3cm}
\end{figure}

Using linear elastic strain energy theory, we can compare the rotational stiffness of each of the beams following:

\begin{equation}
     \kappa = \frac{M}{\theta} 
     \label{eq:shearStress}
\end{equation}

where $\kappa$ is the rotational stiffness, $M$ is the applied moment, and $\theta$ is the angular displacement. The values for the 0.4mm SLL in Table \ref{tab:ssl-analysis} result in a stiffness of 1.9Nmm while the values from Fig. \ref{fig:triangleFEAresults} result in 265.5Nmm for the thirty triangle TRL and 366.6Nmm for the two triangle TRL. The inclusion of triangles to the Strain Limiting Layer has a dramatic impact on torsional resistance. The difference between number of triangles is marginal on bending stiffness relative to the performance improvement without triangles. We choose to design a gripper using the maximal number of triangles tested to reduce stress concentration within the structure. For systems that require the maximum in-plane deformations, the two-triangle design would be better. If maximal resistance to twisting is required, the five-triangle design is the best configuration.

The in-plane deformation value for each configuration was plotted against the corresponding number of triangles ($T_n$) that were present in the Strain Limiting Layer, and the results are shown in Fig.  \ref{fig:triangleFEAresults} (a). Same setup was followed for simulating torsion, and a bending moment of 5Nmm was applied to the free end of the TRL as shown in Fig.  \ref{fig:angle}, and the corresponding results are shown in Fig.  \ref{fig:triangleFEAresults} (b). The angle ($\beta$) between line A and line C was calculated using Eq. \ref{eq:angle-calculation} as shown below.

\begin{equation}
    \beta = \sin^{-1}\bigg(\dfrac{BC}{\sqrt{(AB)^2 + (BC)^2}}\bigg)
    \label{eq:angle-calculation}
\end{equation}

\noindent where, AB is the distance from the center to the edge, BC is the linear displacement in vertical direction, $\beta$ is the angle between line AB and AC.

In-plane deformations reach 20.7mm and 15.6mm for the two and thirty triangles, with a minimum at 11.3mm for the seven triangle case. The torsion test sees values of 0.8$^{\circ}$ and 1.6$^{\circ}$ for two and thirty triangles, with a minimum of 0.58$^{\circ}$ for the five triangle design. The best ratio of in-plane to out-of-plane performance is the two triangle TRL. However, this results in a large stress concentration, limiting cycle life. If a soft deployable structure is being used, the two triangle design should be considered. For the repeated cycles of a soft gripper, we move forward with the thirty triangle design by adding mount points to the structure.

\subsection{Fabrication of TRL Gripper}
\label{fab}
\begin{figure}[t]
    \centering
    \includegraphics[width=0.45\textwidth]{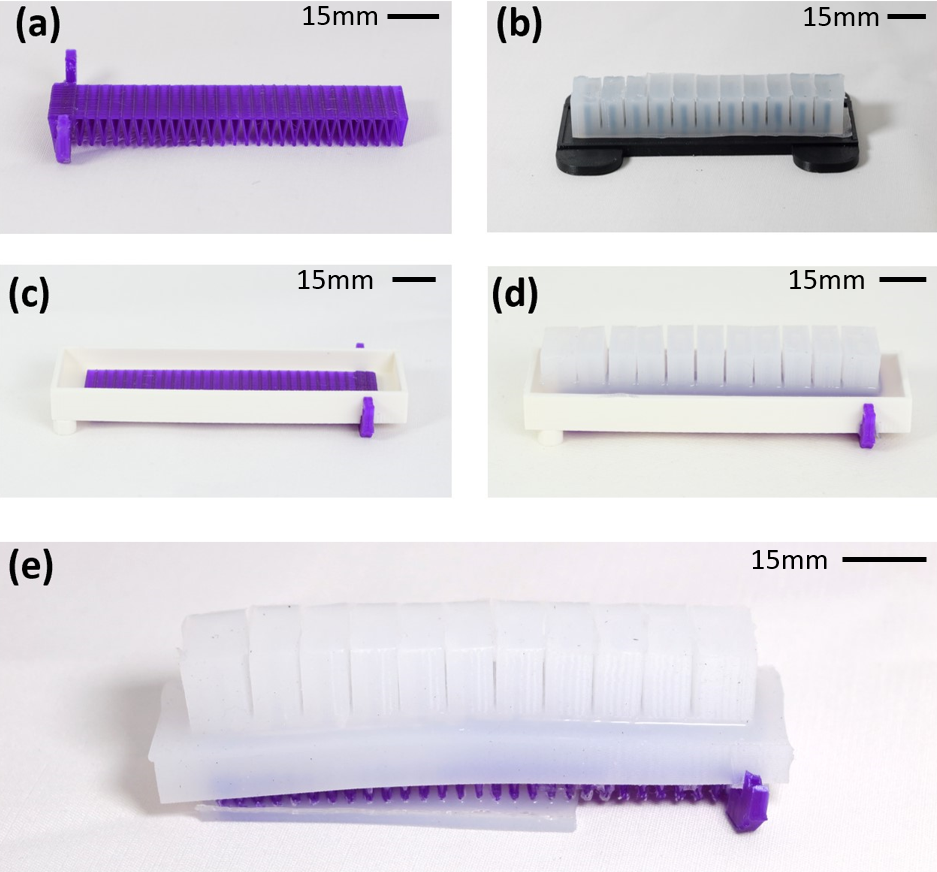}
    \caption{This figure shows the casting process for a TRL gripper. First, (a), we print the 100mm long torsionally resistant Layer (TRL), then in (b), we cast the Pneu-net chambers out of Ecoflex 00-30 following instructions from the Soft Robotics Toolkit. In (c), we insert the TRL into the sacrificial mold and mix Ecoflex. After 25 minutes, we pour until 2mm above the TRL. In (d), we add the Pneu-net chambers and seal them with a thin layer of Ecoflex. Finally in (e), we remove the TRL gripper from the mold, place it spikes side down in a 1mm thick layer of Ecoflex in the Pneu-nets Strain Limiting Layer mold, cure it, and remove.}
    \label{fig:grip-casting}
    \hfill
\end{figure}

The process for the TRL gripper can be seen in Fig. \ref{fig:grip-casting}. The casting process for the TRL gripper and the benchmark gripper follow the process detailed in the Soft Robotics Toolkit \cite{holland2014soft} with the following modifications made. Mixing was done manually following manufacturer instructions. For both grippers, we perform an additional degassing phase after mixing the two Ecoflex 00-30 components until bubble transport stops visually. For the insertion of the TRL, we allow the ecoflex to set for 30 minutes before adding the TRL to ensure the proper set height of 2mm for the TRL.

\section{Results - Experimental Comparison of TRL and Standard Pneu-net} 
In this section, we compare our TRL gripper against the benchmark Pneu-net gripper from the Soft Robotics Toolkit \cite{holland2014soft}. We test both grippers in a bending characterization test. They are tested in plane and out of plane unloaded and with a weight at the distal tip. This simulates the performance of the grippers grabbing an object. We also conduct an extension test on both sets of grippers. They grab a cylinder, which is then pulled out of their grasp. This demonstrates the maximum payload capacity for the grippers and can be used to find the effective stiffness of the grippers and their torsional spring constant. Finally, the TRL gripper is tested on a subset of the YCB dataset. The subset only features objects larger than the maximum payload capacity of a standard Pneu-net gripper. We show success on the test set and only fail on one object within the theoretical payload capacity predicted in Eq. \ref{eq:forceG}.

\subsection{Bend Characterization} 
\begin{figure}[t]
    \centering
    \includegraphics[width=0.45\textwidth]{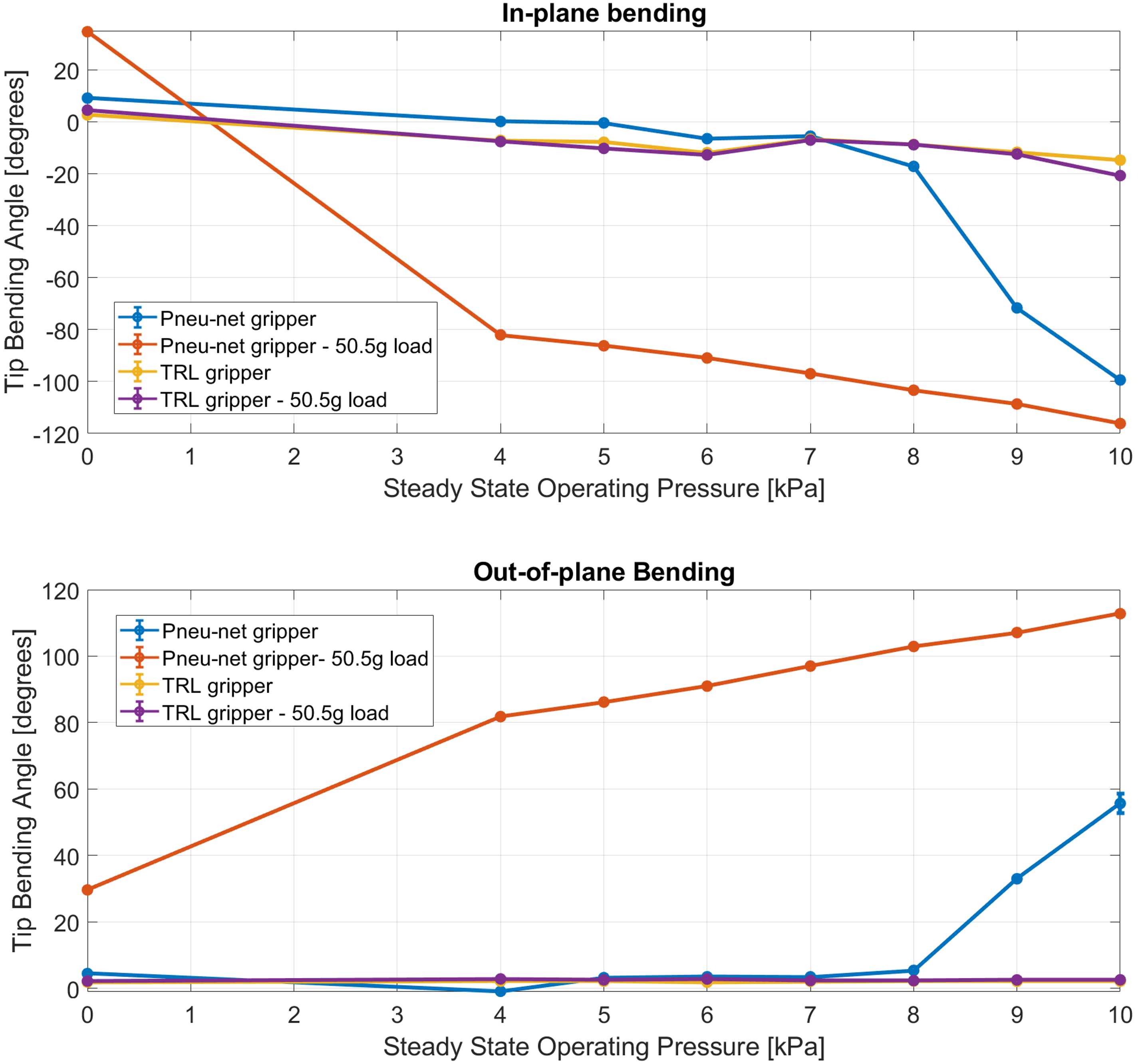}
    \caption{
    Tip bending angle for the TRL and benchmark Pneu-net grippers are shown. Data are presented with and without the addition of a point load. The TRL does an excellent job of reducing out-of-plane deformations. The benchmark gripper showed much larger deformations both in- and out-of-plane. The higher torsional stiffness of the TRL gripper results in better understanding of where picked objects are and in a larger payload capacity.}
    \label{fig:bendingResults}
    
\end{figure}
We evaluate in-plane bending and resistance to torsional deformations for our gripper. Better in-plane performance means that more actuation energy goes into conforming to the object being gripped as less energy is used to deform the structure. Resistance to torsional deformations determines one of the main failure modes for soft grippers. Additionally, smaller torsional deformations can allow a more precise understanding of where grasped objects are. We compare our TRL gripper to the benchmark Pneu-net in terms of bending characterization for both unloaded and loaded tests.

\begin{figure}[t]
    \centering
    \includegraphics[width=0.45\textwidth]{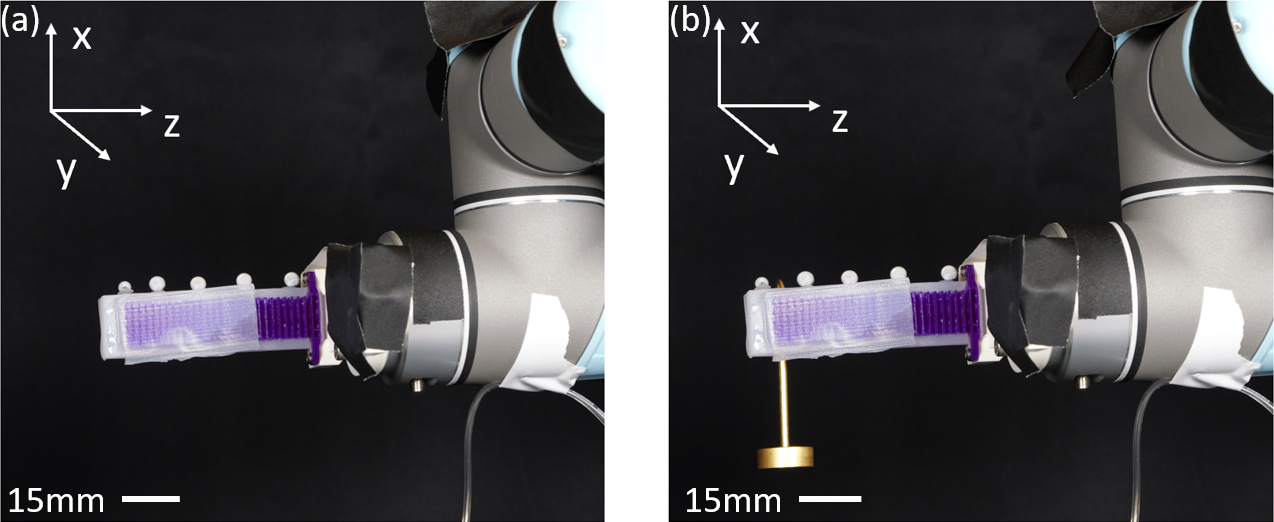}
    \caption{Test setup for Optitrack tip characterization. The TRL gripper is  mounted to the robot arm in a) and shown with TRL a weight of 50 grams in b). The weight was placed between the furthest air chambers, with the mass resting on the silicone forming the SLL. The weight shows how the gripper behaves under loading. We track the deformation of the gripper as steady state pressure is changed using Optitrack markers. This characterization can be used to model the deformation of the TRL gripper as a function of pressure and to understand how lifting an object near the tip of the gripper effects performance. Positive in-plane tip bending is defined in z direction and positive out-of-plane bending is defined in negative Y direction.}
    \label{fig:robot-arm-gripper-trl}
\end{figure}

A series of optical tracking tests were done to determine the tip bending angle of our TRL gripper and the benchmark Pneu-net gripper, as shown in Fig. \ref{fig:robot-arm-gripper-trl}. The XY plane was defined as normal to the robot flange frame, in-plane deformations occurred in the YZ plane, and out-of-plane deformations occurred in the XZ plane. The grippers are evaluated with no load, and with a 50.5g mass positioned between the furthest two air chambers. Pressures ranged from 4kPa to 10kPa and the data can be seen in Fig. \ref{fig:bendingResults}.

We determined bend angles using optical tracking data from markers on the distal tip of the gripper as shown in Fig. \ref{fig:robot-arm-gripper-trl}. In-plane bending was computed in the Y-Z plane, while out of plane bending was computed in the X-Z plane using the inverse tangent function. We recorded data using an OptiTrack motion capture system using Flex 13 cameras. The system was allowed to come to steady state before the data was recorded. The last 120 data points were averaged and a standard deviation determined.

 
Looking at relative in-plane performance of the grippers, we see the TRL gripper do a better job of maintaining its position at rest. This is especially true when the gripper is loaded as the TRL does not meaningfully change tip angle, whereas the benchmark Pneu-net gripper moves more than 30$^{\circ}$ when the weight is added to the distal tip.
 
 

 The unloaded TRL reduced deformations by 57.8\% at rest. When maximally pressurized, the TRL reduced out-of-plane deformations by 96.1\%. When loaded with the 50.5g point mass, the TRL reduces deformations by 92.5\% at rest and by 97.7\% at maximal actuation. This matches closely with our simulated out of plane bending reduction of 94.7\% for the rigid Strain Limiting Layers.
 
\subsection{Grip Slip Force Characterization} 
\begin{figure}[t]
    \centering
    \includegraphics[width=0.45\textwidth]{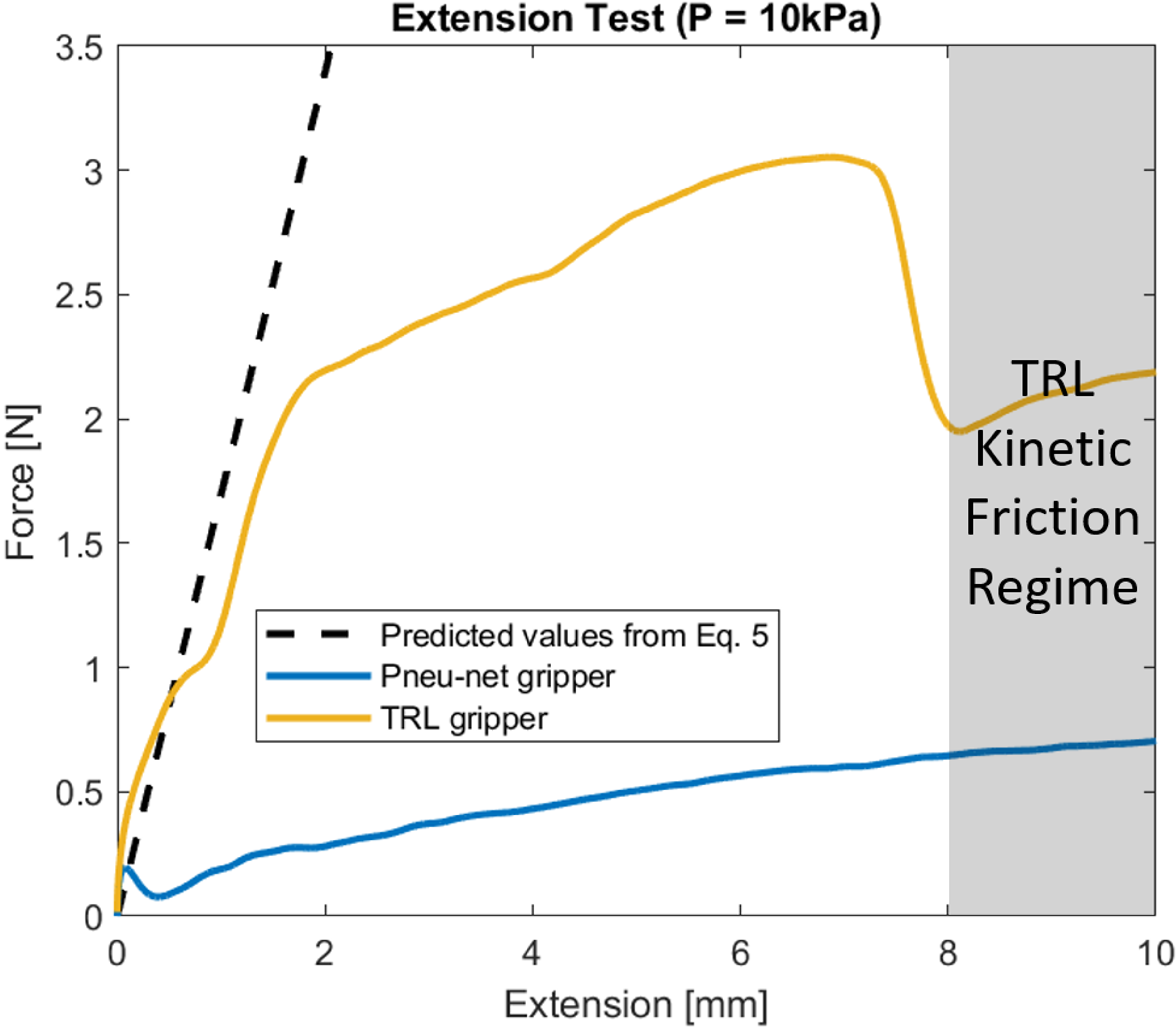}
    \caption{Data from a single extension test shows an increase in maximum force response for the TRL gripper. The TRL gripper demonstrates peaks of 3N grip force, compared to 1N from the benchmark gripper. As seen in our supplemental video, the Pneu-net gripper without a TRL quickly slips due to insufficient torsional stiffness, resulting in a much lower gripper stiffness. At 1.8mm extension, the TRL gripper demonstrates a 7.8x increase in effective grip stiffness. The larger grip stiffness demonstrates a higher minimum antipodal grasp failure condition and can allow for better understanding of an object's position when grasped.}
    \label{fig:instronResults}
    \vspace{-0.5cm}
\end{figure}

\begin{figure}[h]
    \centering
    \includegraphics[width=0.45\textwidth]{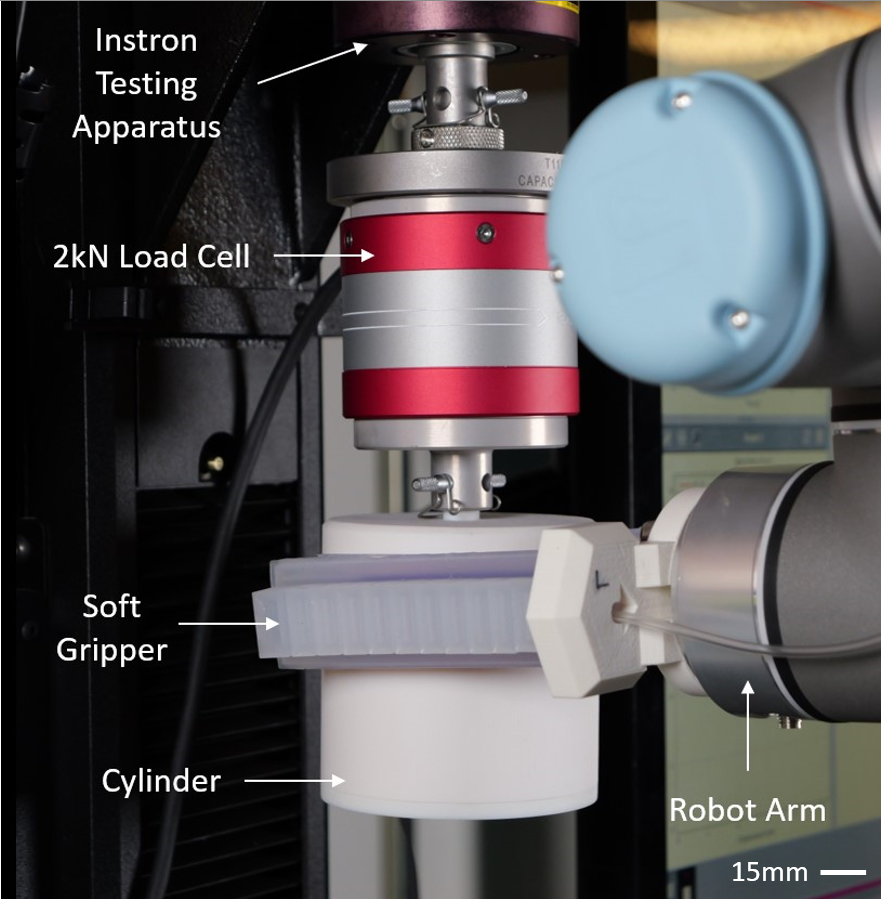}
    \caption{The Instron testing setup for gripper characterization. The two grippers are fixed to a UR5 robot arm. They are positioned to antipodially grasp a 3D Printed cylinder that is 90mm in diameter. The grippers are pressurized to 10 kPa before beginning the test. The Instron testing apparatus moves up and the force response is collected by the transducer. This demonstrates the minimum of the three common antipodal failure modes for soft grippers; slipping, twisting, and shearing.}
    \label{fig:load-capacity}
\end{figure}

We compared the grip strength for a TRL gripper and the benchmark Pneu-net gripper. This test demonstrates the minimum of the three common antipodal failure modes for soft grippers; slipping, twisting, and shearing. Data was gathered using an Instron 68SC-2 at 200Hz as can be seen in Fig. \ref{fig:load-capacity}. The grippers were closed around a 90 mm diameter test cylinder made from VeroWhite on a Stratasys J750 Digital Anatomy printer. The diameter of the cylinder was matched to the distance between antipodal grippers. The grippers were mounted to a Universal Robotics UR-5 and pressurized to 10kPa. The cylinder was moved upwards at a rate of 10mm/s for 100 mm. The data from the experiment can be seen in Fig. \ref{fig:instronResults}. Our gripper shows much larger forces compared to the benchmark gripper reaching 3N compared to 1N. We also demonstrate a 7.8x increase in grasp stiffness. This shows that the TRL gripper has a higher minimum for the three antipodal failure modes, and can lift larger payloads. 

We can see a significant difference in performance of the TRL and standard Pneu-net gripper in Fig. \ref{fig:load-capacity}. A peak force of 3N was observed for the gripper with TRL, while the benchmark gripper demonstrated a peak force of 1N. The TRL gripper has a higher minimum value for the three common failure conditions for an antipodal grasp; slipping, twisting, and shearing. The TRL gripper showed a much higher stiffness compared to the benchmark gripper. Over the first 1.8mm, the benchmark gripper had a stiffness of 152N/m while the TRL gripper showed a stiffness of 1181N/m. By using the TRL instead of the benchmark SLL, we demonstrate a 7.8x increase in grasp stiffness. A larger grip stiffness demonstrates a higher minimum antipodal grasp failure condition and can allow for better understanding of where objects are within the grasp.

\subsection{YCB Object Grasping}
\begin{figure} [t]
    \centering
    \includegraphics[width=0.45\textwidth]
    {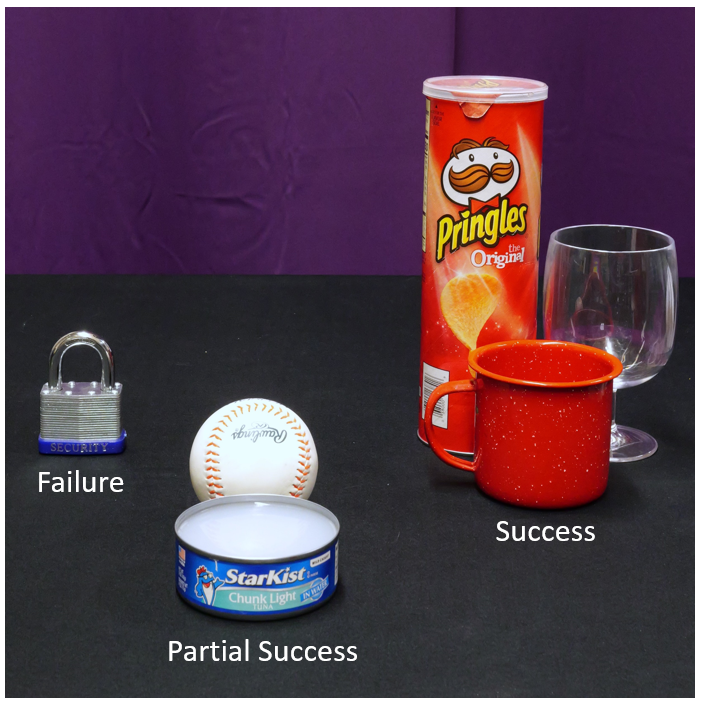}
    \caption{This figure shows the YCB objects we evaluated the TRL gripper against. All of the items were chosen to be above the maximum demonstrated capacity for the Pneu-net gripper so any objects grasped show added capability. Additionally we filtered to non-box objects that fit within the jaws of the TRL gripper. Of these items, the TRL lifted five items successfully, with two being partial successes. The only failure is the padlock.}
    \label{fig:YCB Objects}
\end{figure}

We demonstrate the TRL gripper's ability to pickup objects from the YCB object dataset \cite{YCBObjects}. We chose to limit our object selection to non-box objects larger than 100g and smaller than 300g. This was set as 100g is the maximum possible lifting force applied by the benchmark gripper and 300g was the limit for the TRL gripper. Boxes were excluded due to their lack of compliance required for their grasp. From this set, we filtered out objects that did not fit between the TRL grippers using the mount shown in Fig. \ref{fig:load-capacity}. In total this left six objects in our test set as seen in Fig. \ref{fig:YCB Objects}. All objects were filled to match their weight and approximate mass distribution according to the data from \cite{YCBObjects}.

We were able to successfully lift five of the objects, with two of the five being partial lifts. A successful lift is fully lifting the object off the ground and maintaining contact. A partial success is lifting the object but not maintaining contact. The objects the TRL gripper successfully grasped were the Pringles can, the wine glass, and the mug. The partial successes were the tuna fish can and the baseball. The baseball slipped out of the grasp due to pitching backward. The tuna fish can slipped out due to roll. The TRL gripper was unable to grasp the padlock due to it being too narrow to generate enough normal force. If the gripper mount were closer together, the TRL gripper would have likely been able to lift the padlock. 

We demonstrated a significant increase in capability for the TRL gripper compared to the Pneu-net gripper. Not only did we show the addition of the TRL increased normal force and torsional stiffness, but it translated to a greater capability set for Pneu-net grippers. This can enable their use on a wider variety of tasks and reduces a key limitation of their adoption.

\section{Conclusion and Future Work}


In this paper, we developed a Torsion Resistant Strain Limiting Layer (TRL) to increase payload capacity by increasing torsional resistance. A comprehensive design study was conducted using FEA simulations to understand the design landscape of triangulated beams. The number of triangles on a 100mm-long Strain Limiting Layer was varied to characterize the performance of the TRL in plane and out of plane. It was found that the overall change in angular displacement was less than two degrees for all cases. However, the TRL with thirty triangles showed good in-plane bending characteristics and evenly distributed stresses along the length of the structure. A modified casting process was adopted to integrate the TRL to a standard Pneu-net actuator, and extension and Optitrack tests were conducted to see the performance improvement over the standard Pneu-net actuator. 

This work found that adding a Torsion Resistant Strain Limiting Layer dramatically increased torsional performance, reducing deformations by up to 97.7\%. The addition of the TRL allowed a standard Pneu-net gripper to pick up a 5kg dumbbell, maxing out the capacity of the robot arm it was installed on. The TRL gripper was also tested on a subset of the YCB data set and successfully picked up all but one item, all of which were heavier than the maximum demonstrated payload capacity for a Pneu-net gripper without the TRL. The TRL gripper also demonstrated increased the grasp stiffness of 7.8x the standard Pneu-net. Future work for these TRL grippers include sensorization and co-optimization of cast material properties with TRL materials. Additionally, adding a variable gap between the TRL gripping surfaces would allow them to lift a larger variety of objects. This work lays the foundation to easily incorporate TRLs into existing soft robots and demonstrates additional payload capacity that is often lost due to torsional deflection in soft grippers.

\bibliographystyle{IEEEtran}
\bibliography{hsa}
\end{document}